\documentclass[sigconf]{acmart}
\AtBeginDocument{%
  }

\copyrightyear{2026}
\acmYear{2026}
\setcopyright{cc}
\setcctype{by}
\acmConference[DAC '26]{63rd ACM/IEEE Design Automation Conference}{July 26--29, 2026}{Long Beach, CA, USA}
\acmBooktitle{63rd ACM/IEEE Design Automation Conference (DAC '26), July 26--29, 2026, Long Beach, CA, USA}
\acmDOI{10.1145/3770743.3804346}
\acmISBN{979-8-4007-2254-7/2026/07}

\usepackage{cite}
\usepackage{url}
\usepackage{amsmath,amsfonts}
\usepackage{algpseudocode}
\usepackage{graphicx}
\usepackage{textcomp}
\usepackage{xcolor}
\usepackage{xspace}
\usepackage{comment}
\usepackage{kotex}
\usepackage{enumitem}
\usepackage{booktabs}
\usepackage{multirow}
\usepackage{makecell}
\usepackage[normalem]{ulem}
\usepackage{algorithm}
\usepackage{algpseudocode}
\usepackage{indentfirst}
\newcommand{\SysName}{ORBIS\xspace}
\title{\SysName: Output-Guided Token Reduction with Distribution-Aware Matching for Video Diffusion Acceleration}
\author{Hangyeol Lee}
\affiliation{%
  \institution{KAIST}
  \city{Daejeon}
  \country{South Korea}}
\email{lhg4294@kaist.ac.kr}
\author{Joo-Young Kim}
\affiliation{%
  \institution{KAIST}
  \city{Daejeon}
  \country{South Korea}}
\email{jooyoung1203@kaist.ac.kr}
\authornote{Corresponding author.}

\begin{document}
\begin{abstract}
Diffusion Transformer (DiT) has emerged as a powerful model architecture for generating high-quality images and videos. In the case of video DiT, 3D Spatio-Temporal Attention increases token length in proportion to the number of frames, sharply increasing computational cost. Token reduction methods mitigate this cost by exploiting spatial redundancy, but existing approaches rely on inaccurate similarity estimates and lightweight matching algorithms, resulting in poor matching quality and only marginal acceleration.

To overcome these limitations, we propose ORBIS, an SW–HW co-designed accelerator for video DiT. ORBIS leverages the output activation from the previous timestep to obtain more accurate inter-token similarity, substantially improving matching quality and enabling a higher token reduction ratio. We further introduce a Distribution-Aware Token Matching (DATM) algorithm that captures global token distribution and explicitly minimizes token-pair loss for additional gains.
To fully hide DATM latency, we design specialized, deeply pipelined hardware and minimize its hardware cost through quantization, occupying only 2.4\% of total area with negligible accuracy loss. Extensive experiments show that ORBIS achieves about 2$\times$ higher token reduction ratio than the state-of-the-art approach, AsymRnR, while delivering up to 4.5$\times$ speedup and 79.3\% energy reduction compared to an NVIDIA A100 GPU.
\end{abstract}

\maketitle
\vspace{-0.05in}
\section{Introduction}

\label{section1}

\begin{figure} [t]
  \centering
  \includegraphics[width=\linewidth]{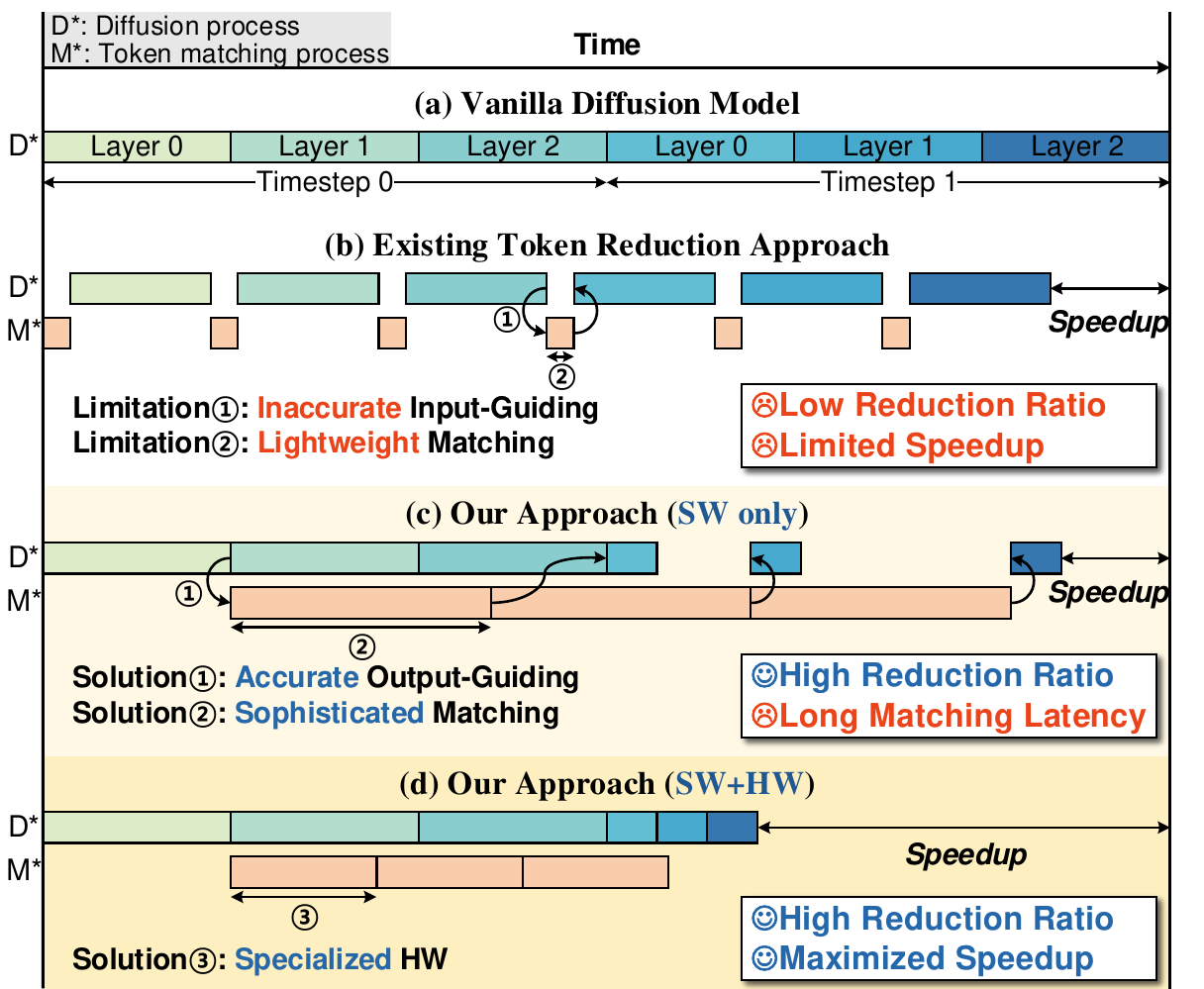}
  \vspace{-0.3in}
  \caption{Execution timeline of four scenarios. For illustration, the diffusion process is simplified to two timesteps with three layers each.}

  \label{fig1}
  \vspace{-0.26in}
\end{figure}

Recent progress in image~\citep{rombach2022high, peebles2023scalable, chen2024pixart} and video generation~\citep{hunyuanvideo2024, yang2024cogvideox, wan2025wan22, genmo2024mochi, yuangroup2024opensoraplan} has been primarily driven by innovations in diffusion models, which generate samples through an iterative denoising process over multiple timesteps. Building on this foundation, Diffusion Transformer (DiT)~\citep{peebles2023scalable} backbone further advances performance by effectively modeling global dependencies and scaling efficiently with model size. In video DiT, the model adopts 3D Spatio-Temporal Attention to extend image DiT into the temporal domain, ensuring temporal coherence. However, the introduction of 3D Attention substantially increases computational cost, as the token length scales proportionally with the number of frames. Combined with the quadratic complexity of the attention mechanism, this scaling substantially increases the computational workload of video DiT, resulting in substantial energy use and prolonged inference times. For instance, generating a 129-frame 544p video with the HunyuanVideo model~\citep{hunyuanvideo2024}  takes over 0.5 hours on an NVIDIA A100 GPU~\citep{nvidiaA100}, showing the heavy computational burden of video DiT.

To address the computational challenges of video DiT, several acceleration approaches have been proposed. Distillation-based approaches~\citep{ meng2023distillation, sauer2024add, luo2023lcmlora} aim to reduce the number of sampling steps and overall network complexity. However, these approaches typically require extensive retraining, resulting in significant training costs. Meanwhile, training-free caching approaches~\citep{selvaraju2024fora, zhao2025pab, chen2024deltadit, liu2025tgate, kahatapitiya2024adacache, lv2025fastercache, zou2025toca, liu2025astraea} skip redundant computations by reusing activations from previous timesteps, leveraging the temporal redundancy inherent in the diffusion model. However, these methods suffer from high memory demand, as all data to be reused for subsequent timesteps must be stored, and this burden becomes more severe in video DiT, where the number of tokens and activation size increase proportionally with the number of frames.
For example, applying ToCa~\citep{zou2025toca} or AdaCache~\citep{kahatapitiya2024adacache} to the HunyuanVideo model at 129 frames and 544p incurs an additional 77 GB of memory. Given a base requirement of 45 GB, this results in a substantial total of approximately 122 GB, which is about 2.7 $\times$ the baseline. Moreover, this memory overhead is expected to exacerbate with model and spatiotemporal scaling, posing a significant bottleneck for future large-scale video DiTs.

In contrast to caching approaches, token reduction approaches~\citep{bolya2023tomesd,sun2025asymrnr,lu2025toma} reduce computational cost by leveraging spatial redundancy among tokens within each timestep. Notably, these approaches incur only negligible memory overhead, since redundancy is represented by compact token index pairs rather than full activation tensors.
% 너무 low level임
These token index pairs are the core of the token reduction approach, as they determine which tokens are reduced and how the reduced tokens are reconstructed, thereby directly tied to its overall performance. However, prior studies have struggled to effectively determine these token index pairs, resulting in only marginal improvements in practical acceleration.

As illustrated in Fig.~\ref{fig1} (b), existing token reduction approaches suffer from two fundamental limitations that ultimately degrade their overall performance. 
\textbf{(1)} They perform token matching based on the layer’s input activation (input-guiding). However, this activation provides only an inherently inaccurate approximation of the ground-truth output-token similarity, resulting in poor matching quality, leading to low reduction ratios and consequently limited speedup.
\textbf{(2)} This input-guided matching forces these methods to adopt lightweight token matching algorithms. 
However, such lightweight designs incur substantial token-pair losses and further degrade matching quality. 
This constraint arises because matching must be initiated immediately before the layer where token reduction is applied. As it cannot overlap with the diffusion process, its latency becomes fully exposed, and only low-latency lightweight algorithms remain viable.
As a result, these two limitations jointly degrade the matching quality, forcing the use of low reduction ratios, which in turn leads to limited speedup.

To address these two limitations, we propose ORBIS, which consists of three main designs: Output-Guided Matching (OGM) for more accurate token similarity estimation, Distribution-Aware Token Matching (DATM) that explicitly minimizes token-pair loss, and a specialized hardware accelerator for fully hiding the execution latency of the DATM pipeline.

OGM leverages the previous timestep’s output activation of the target layer for token matching. This provides a much closer approximation of the true output-token similarity than input-guided matching, consequently achieving a higher reduction ratio. Moreover, this output-guided design also allows the matching process to be initiated in the previous timestep, partially hiding its latency, making it viable to employ more sophisticated matching algorithms. Building on this opportunity, we propose DATM, which accounts for the global token distribution and progressively minimizes token-pair loss, thereby delivering additional performance gains. As shown in Fig.~\ref{fig1}(c), the combination of OGM and DATM achieves a significantly higher token reduction ratio compared to Fig.~\ref{fig1}(b). However, although OGM allows the matching process to overlap with the diffusion process, the matching latency of DATM remains high because it requires repeated large-scale vector processing for loss computation. As a result, its latency cannot be fully hidden, which in turn limits the attainable speedup. To address this, we design specialized hardware with a deeply pipelined datapath tailored to DATM’s vector-processing pattern. In addition, we apply quantization to significantly reduce the hardware cost with negligible impact on accuracy. This hardware fully hides the latency of DATM and maximizes the attainable speedup (Fig.~\ref{fig1} (d)).

Our main contributions are summarized as follows:
\begin{itemize}
    \item We are the first to propose an output-guided token reduction approach, which provides more accurate token similarity estimation, thereby enabling a higher reduction ratio.
    \item We propose a Distribution-Aware Token Matching (DATM) algorithm that jointly considers global token distribution and explicitly minimizes token-pair loss, yielding additional performance improvements.
    \item We develop a specialized hardware accelerator for the DATM algorithm that introduces only minimal hardware overhead, while fully hiding its latency and thereby maximizing the attainable speedup.
    \item ORBIS achieves approximately 2$\times$ higher token reduction ratio than the prior software-based method AsymRnR~\citep{sun2025asymrnr}, while also delivering up to 4.5$\times$ speedup and  79.3\% energy reduction compared to an NVIDIA A100 GPU.
\end{itemize}
\section{Background \& Motivation}
\label{sec:2}

\subsection{Diffusion Model and Token Reduction Approach}
\label{sec:2A}

\begin{figure}[t]
  \centering
  \includegraphics[width=\linewidth]{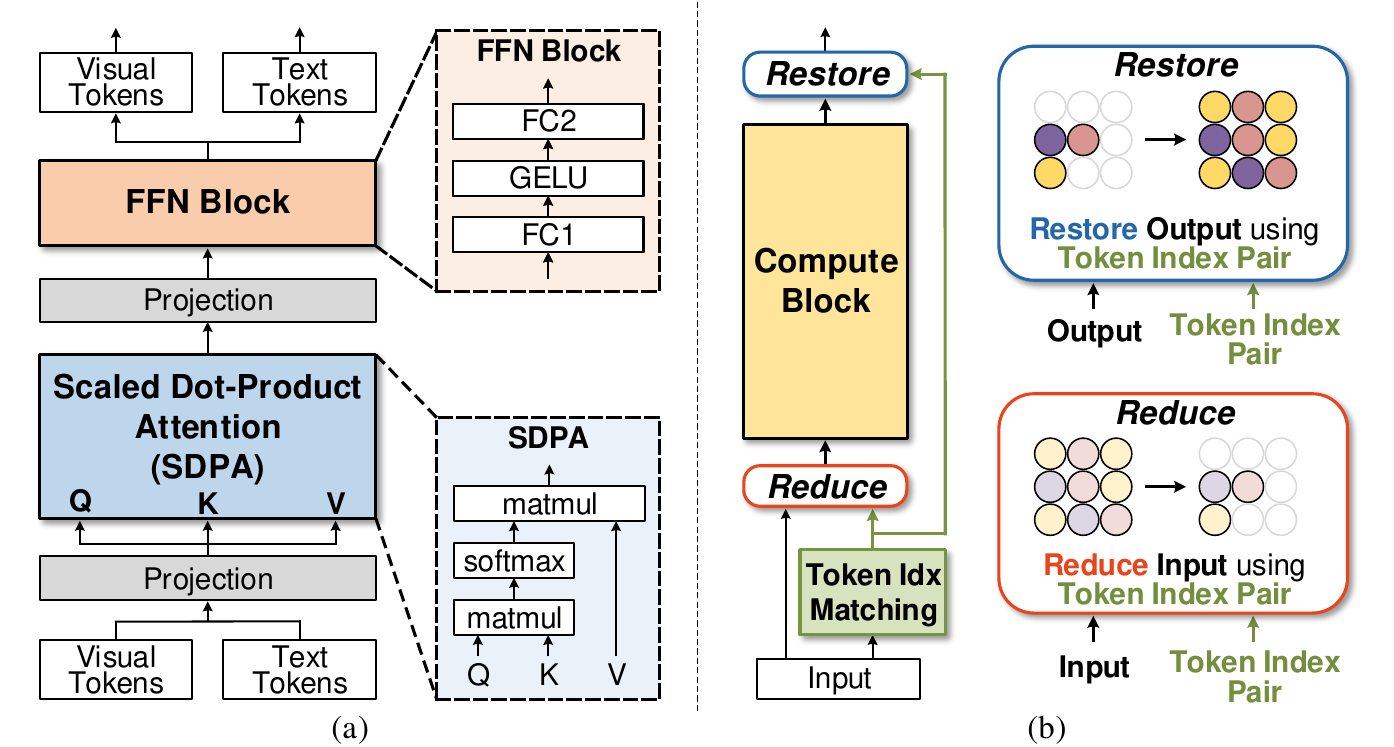}
  \vspace{-0.25in}
  \caption{(a) The structure of MMDiT block. (b) The process of the token reduction approach}
    \vspace{-0.2in}
  \label{fig2}
\end{figure}
In diffusion models, the generation process begins with pure noise and progressively removes it over multiple denoising steps to produce the target sample (e.g., images or videos). Recently, DiT has become the predominant backbone for the denoising network due to its scalability and high-quality output. Building on DiT, modern diffusion models adopt Multi-Modal DiT (MMDiT), replacing the traditional separation between self-attention and cross-attention with unified attention over a concatenated token sequence. As shown in Fig.~\ref{fig2} (a), visual and text tokens are concatenated and jointly projected into Q, K, and V, then processed by a Scaled Dot-Product Attention (SDPA) followed by a Feed-Forward Network (FFN). These MMDiT blocks are stacked to form the denoising network, which is executed repeatedly over dozens of timesteps and thus incurs substantial computational cost.

The token reduction approach is an effective way to reduce computational cost by exploiting spatial redundancy among tokens within each timestep. Unlike caching methods, which store activations for reuse across timesteps, token reduction records only token index pairs, incurring negligible memory overhead.
Fig~\ref{fig2} (b) illustrates the pipeline of existing token reduction approaches~\citep{bolya2023tomesd,sun2025asymrnr}, which proceeds as follows:
\begin{enumerate}
\item \textbf{Token Index Matching:} Given the input tensor, perform token matching based on its inter-token similarity to generate token index pairs.
\item \textbf{Reduce:} Apply the pairs to shorten the token sequence.
\item \textbf{Compute:} Process the reduced tokens with the target compute block (e.g., SDPA, FFN, etc.).
\item \textbf{Restore:} Apply the pairs to reconstruct the full-length output from the reduced result.
\end{enumerate}

% These token index pairs, used by the Reduce and Restore processes, determine which tokens are reduced and how they are reconstructed, thereby directly tied to the overall performance of token reduction approach.
These token index pairs play a central role in the Reduce and Restore processes, specifying which tokens are reduced and how the reduced tokens are reconstructed. Thus, their quality directly impacts the overall performance of token reduction methods. In the following sections, we discuss two key factors that degrade this quality and consequently limit the overall performance.
\begin{figure}[t]
  \centering
  \includegraphics[width=\linewidth]{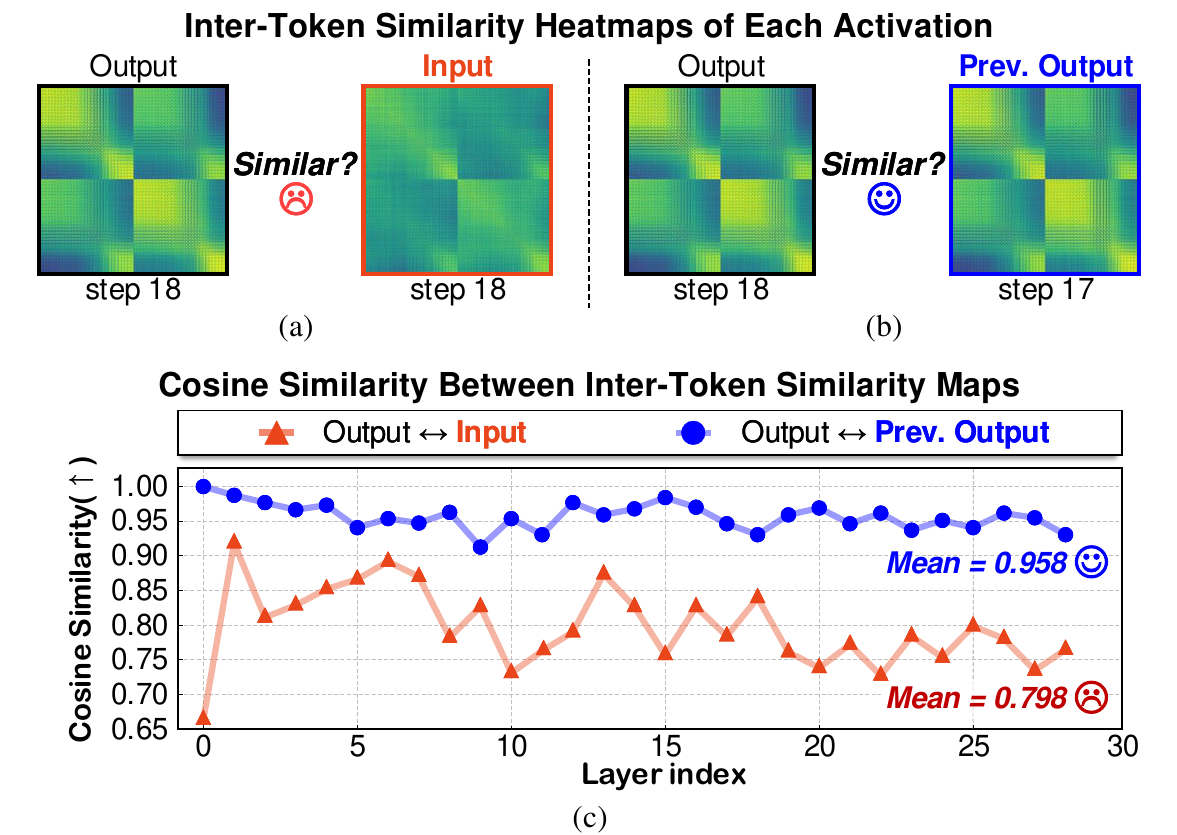}
  \vspace{-0.25in}
  \caption{Inter-token similarity heatmaps of the output activation versus (a) Input and (b) Previous output. (c) Cosine similarity between the output inter-token similarity map and those of the input and previous output.}
  \label{fig3}
  \vspace{-0.2in}
\end{figure}

% \vspace{-0.1in}
\subsection{Limitations of Existing Input-Guided Approach}
\label{sec:2B}

Conventional input-guided matching fails to approximate the output inter-token similarity, which is the ground-truth input required for accurate token index pair construction.
The ideal way to construct token index pairs is to use the inter-token similarity within the output activation, because token reduction inherently aims to skip redundant computations across spatially similar output tokens. However, since this ground-truth (GT) output activation is unavailable before processing, the pairs must instead be constructed using a substitute that closely approximates its inter-token similarity. 
% As shown in Fig.~\ref{fig2} (b), prior studies have mainly used the input activation as this substitute input. However, as shown in the heatmap of Fig.~\ref{fig3} (a), the inter-token similarity of the input activation differs significantly from that of the GT output activation.
As shown in Fig.~\ref{fig2} (b), prior studies have mainly used the input activation as this substitute input, but the heatmap in Fig.~\ref{fig3} (a) shows that its inter-token similarity differs significantly from that of the GT output activation.

We observe strong temporal consistency in output activations across adjacent timesteps, and hypothesize their similarity maps are likewise consistent. As shown in Fig.~\ref{fig3} (b), the similarity map of the previous output closely resembles that of the GT output. This trend is further confirmed by the cosine similarity between the GT output similarity map and those derived from the input and previous output across all layers (Fig.~\ref{fig3} (c)). Consistent with the visualization results, the previous output similarity achieved a high correlation of 0.953, far exceeding the input's 0.782. Based on this, we propose an output-guided token matching approach that leverages the previous timestep’s output similarity to predict the current timestep’s output similarity, thereby overcoming the limitations of input-guided matching and improving matching quality, which in turn leads to higher overall performance.

\begin{figure} [t]
  \centering
  \includegraphics[width=\linewidth]{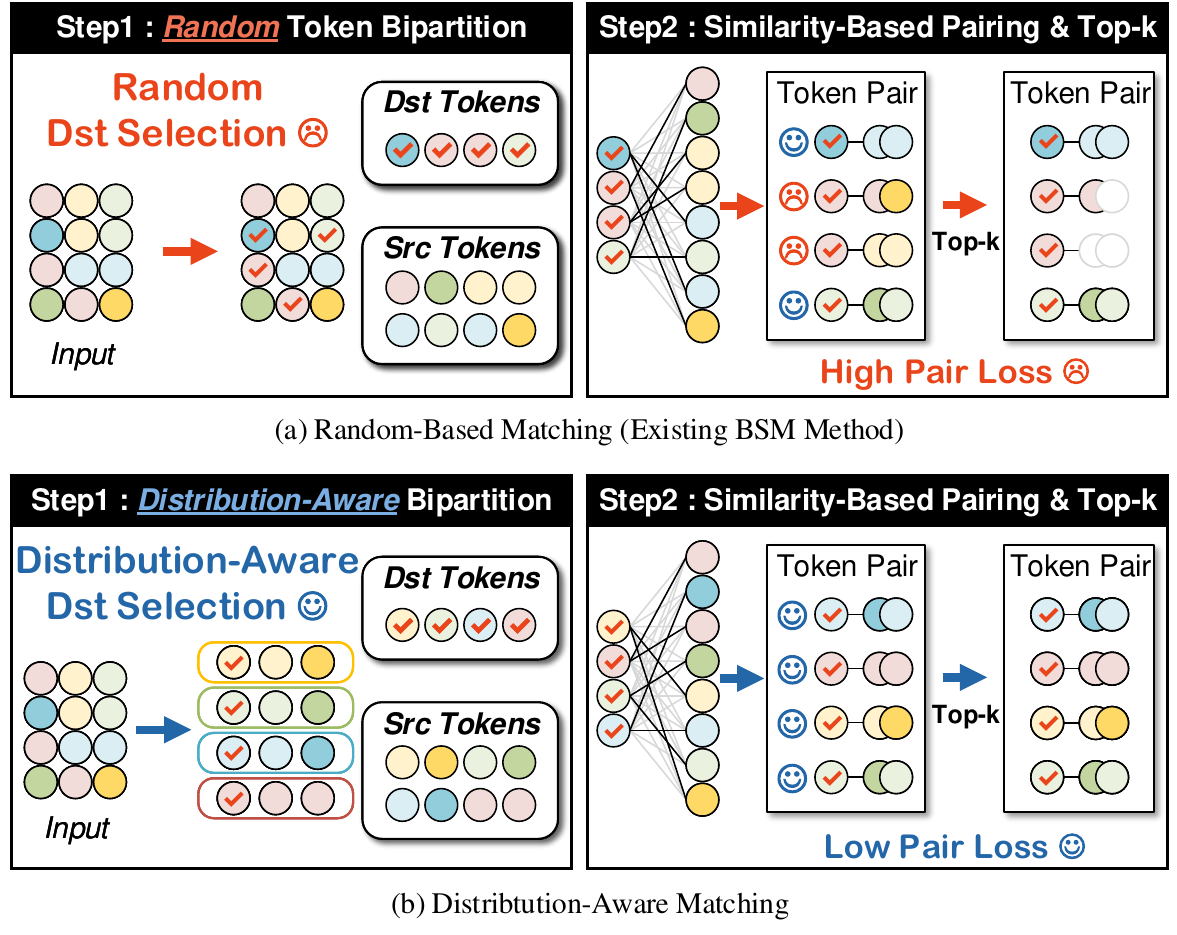}
  \vspace{-0.3in}
  \caption{Token matching algorithm: (a) Random-based. (b) Distribution-aware.}
  % \caption{wip}
  \label{fig4}
  \vspace{-0.3in}
\end{figure}

\vspace{-0.1in}

\subsection{Limitations of Existing Matching Algorithm}
\label{sec:2C}

Existing token reduction often relies on lightweight matching algorithms to minimize overhead, but such lightweight designs inevitably incur large matching losses and degrade the quality of the resulting token pairs. This reliance on lightweight matching arises from the structural limitation of input-guided matching: because it uses the input activation as the matching input, the matching process must be executed immediately before the reduce block. Since this matching occurs in a non-overlappable, pre-processing stage, its latency directly accrues to the end-to-end inference time and cannot be hidden by parallelism. A representative lightweight method is Bipartite Soft Matching (BSM)~\citep{bolya2023tomesd}, shown in Fig.~\ref{fig4} (a), which randomly partitions tokens into destination \textit{(dst)} and source \textit{(src)} sets and then forms token pairs based on similarity, retaining only the top-k pairs according to the predefined reduction ratio.

However, BSM inherently leads to large token-pair losses as its randomly selected \textit{dst} tokens often fail to represent the overall token distribution. For high accuracy, the \textit{dst} tokens must be representative so that each \textit{src} token can find an appropriate reference while keeping the token-pair loss small. However, BSM selects \textit{dst} tokens without considering the global distribution; thus, the references are frequently mismatched, the pair loss is not explicitly controlled, and the resulting pairs are inaccurate and unreliable. As shown in Fig.~\ref{fig4} (a), random selection may exclude a proper reference, forcing a yellow \textit{src} to pair with a red \textit{dst} and incurring large loss. In contrast, Fig.~\ref{fig4} (b) shows that selecting \textit{dst} tokens to reflect the global distribution enables closer matches, yielding more uniform coverage and substantially smaller pair loss. Motivated by this, we introduce a distribution-aware matching algorithm that accounts for the global token distribution and progressively minimizes the token-pair loss, further improving overall performance.

\section{ORBIS’s SW-HW Co-Design Strategy}
\subsection{Output-Guided Token Reduction Approach}
\label{sec:ctp}

\begin{figure}[t]
  \centering
  \includegraphics[width=\linewidth]{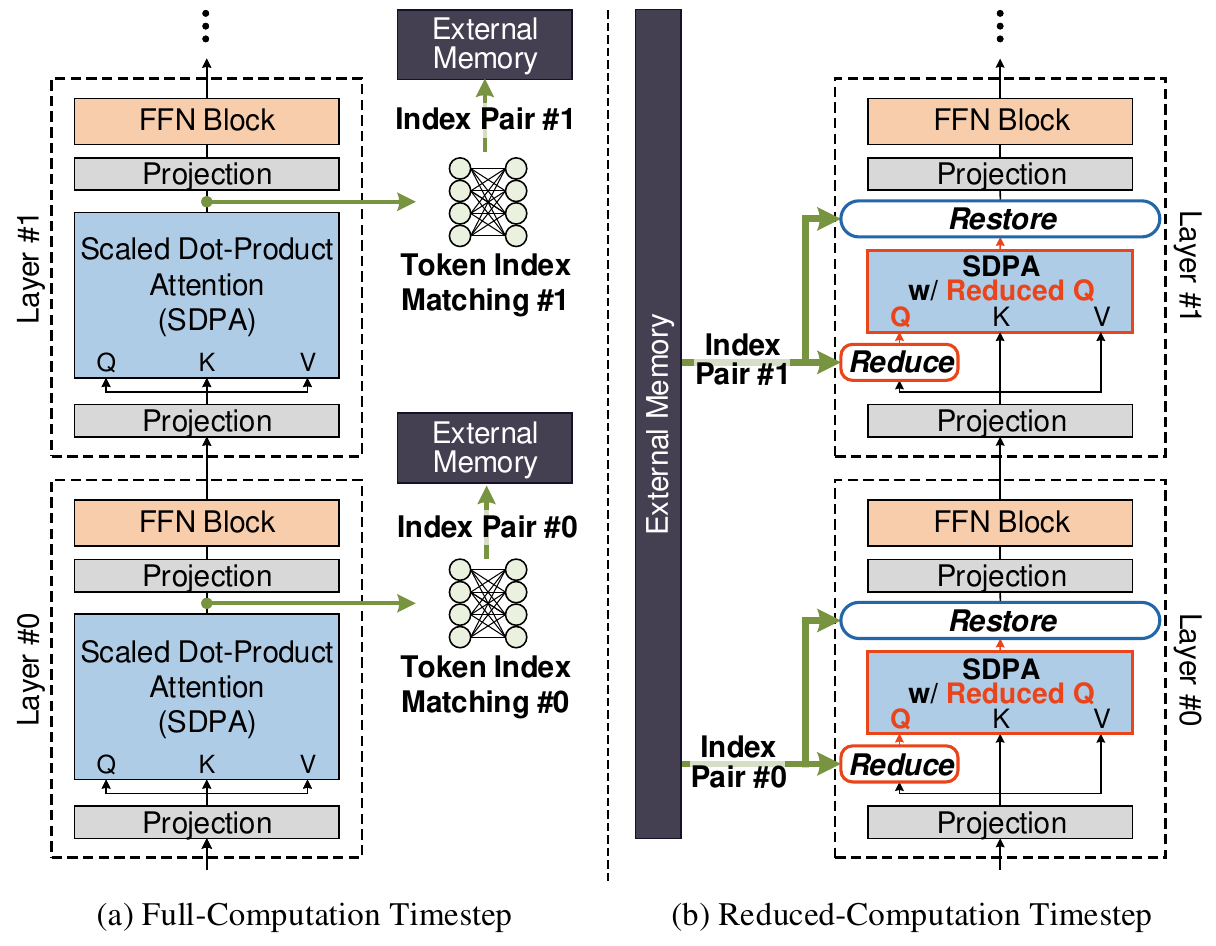}
  \vspace{-0.25in}
  \caption{ORBIS’s diffusion process. Each denoising timestep is divided into (a) Full-Computation and (b) Reduced-Computation timesteps.}
  % \caption{wip}
  \label{fig5}
  \vspace{-0.20in}
\end{figure}

Based on the observation in Sec.~\ref{sec:2B}, we propose an output-guided token reduction approach that performs token matching using the previous timestep’s output activation, which significantly improves matching quality and overall performance. Accordingly, we divide the denoising timesteps into Full Computation (FC) steps and Reduced Computation (RC) steps. As illustrated in Fig.~\ref{fig5}, during FC timesteps, the model performs the main diffusion process and, in parallel, executes token index matching to generate index pairs for the upcoming RC timesteps, where these pairs are immediately used to apply token reduction. Given that Scaled Dot-Product Attention (SDPA) accounts for the majority of the total computation in video DiT, we apply token reduction only to SDPA.

In the FC timesteps, the original DiT layers run without token reduction, while token index matching is performed in parallel at every layer using the SDPA output activations to generate index pairs. These pairs are stored in external memory and later consumed in the subsequent RC timesteps to apply token reduction. Because the matching results are used in the RC timesteps, the matching process can overlap with the main diffusion process in the FC timestep, allowing its latency to be hidden. 
% This is fundamentally different from input-guided approaches, where matching must occur immediately before the reduce block and therefore cannot overlap with the main diffusion process.

On the other hand, in the RC timesteps, token reduction is applied by inserting additional reduce and restore blocks before and after SDPA, respectively. The index pairs generated during the FC timestep are fetched from the external memory and directly used by these blocks. 
Although this design superficially resembles caching-based approaches in that it preserves information from the previous timestep for the next, our method stores only compact index metadata rather than full activation tensors, resulting in negligible memory overhead. In practice, we schedule multiple RC timesteps after each FC timestep and reuse the identical index pairs across a short sequence of RC steps. Empirically, reusing the pairs for three RC timesteps provided the best trade-off, and we adopt this configuration throughout our design.

In summary, by exploiting the temporal consistency\textemdash using the previous timestep’s output activation as the input of token matching\textemdash our design fully leverages the inherent properties of the diffusion process. This yields far more accurate matching than input-guided approaches, enabling more aggressive token reduction while maintaining accuracy and ultimately delivering higher overall performance.
This output-guided design stands in clear contrast to existing input-guided approaches, where matching must occur immediately before the compute block, fully exposing its latency. In our approach, the matching process can overlap with the FC timestep, allowing its latency to be partially hidden and creating headroom to employ more sophisticated matching algorithms. Building on this opportunity, the next section introduces our distribution-aware token matching algorithm, which further improves the matching quality and, in turn, delivers additional performance gains.

\vspace{-0.1in}
\subsection{Distribution-Aware Token Matching Algorithm}
\label{subsec:token_matching_algorithm}
% The conventional token index matching uses input activations as a 
In Sec.~\ref{sec:2C}, we identified that existing random-based token matching leads to large token-pair loss as it ignores the global token distribution and often selects unrepresentative destination (\textit{dst}) tokens. To address this limitation, we propose a Distribution-Aware Token Matching (DATM) algorithm that explicitly incorporates global distribution information by clustering similar tokens and selecting representative tokens in each cluster as \textit{dst} tokens. Specifically, DATM updates the \textit{dst} tokens to explicitly minimize the global token-pair loss and performs this optimization iteratively to continuously refine the token correspondence. Through this loss-driven refinement, DATM significantly improves the quality of the resulting token index pairs, leading to overall performance.

\begin{algorithm}[t]
% \small
\footnotesize
% \scriptsize
\setlength{\baselineskip}{0.1\baselineskip}
\caption{Distribution-Aware Token Matching Algorithm}
\label{alg:otr_quant_kmeans}
\begin{algorithmic}  % [1] → line numbering
\State \textbf{Input:} Input tensor $\mathbf{X}^{N\times D}$; 
       Number of Dst tokens $K$;
\State Convergence pair-loss threshold $\varepsilon$;
       Top-k ratio $r\in(0,1]$
\State \textbf{Output:} Dst--Src token index pairs $\mathcal{P}=\{(i_{\text{dst}}, j_{\text{src}})\}$
\State
\State
\State
\State
\State
\State 
\State \textbf{Step-1) Dst\&Src Initialization.}
\State $\mathbf{Dst}^{K\times D}  \gets \mathrm{RandomSample}(\mathbf{X}, K)$
\State $\mathbf{Src}^{(N-K)\times D} \gets \mathrm{RemainingTokens}(\mathbf{X}, \mathbf{Dst})$
\State
\State
\State
\State
\State
\State
\State \textbf{repeat}
\State \hspace{1em} \textbf{Step-2) Dst-Src Pairing.}
\State \hspace{1em} $\text{Distance}[K,N-K] \gets \mathrm{L2loss}(\mathbf{Dst}, \mathbf{Src})$
\State \hspace{1em} $\text{ClosestDist}[N-K],\; \text{ClosestIdx}[N-K]        \gets \min_{k} \text{Distance}[k, :]$
\State \hspace{1em} $\text{IdxPairs} \gets \{(\text{ClosestIdx}[n],\,n)\mid n=0..N-K-1\}$
\State
\State
\State
\State
\State
\State
\State \hspace{1em} \textbf{Step-3) Convergence Check.}
\State \hspace{1em} $\text{Loss} \gets \mathrm{Average}(\text{ClosestDist})$\Comment{Loss : Token-Pair Loss}
\State \hspace{1em} $\text{if } (\text{PrevLoss}-\text{Loss}  < \varepsilon) \text{ then \textbf{break}}$
\State \hspace{1em} $\text{PrevLoss} \gets \text{Loss}$
\State
\State
\State
\State
\State
\State
\State \hspace{1em} \textbf{Step-4) Dst Update.}
\State \hspace{1em} $\mathbf{UpdatedDst[k]}\gets\mathrm{Average}\{\mathbf{X[n]}\mid(k,n)\in\text{IdxPairs}\},\;k=0..K-1$
\State \hspace{1em} $\mathbf{UpdatedDst} \gets \mathrm{NearestToken}(\mathbf{X},\mathbf{UpdatedDst})$
\State \hspace{1em} $\mathbf{Dst} \gets \mathbf{UpdatedDst}$
\State \hspace{1em} $\mathbf{Src} \gets \mathrm{RemainingTokens}(\mathbf{X}, \mathbf{Dst})$
\State
\State
\State
\State
\State
\State
\State \textbf{Step-5) Top-k Selection.}
\State $\mathcal{P}\gets \mathrm{Topk}( \text{IdxPairs},\text{ClosestDist},r )$\Comment{$\mathcal{P}$ : Final Dst-Src Token Indx Pair}

\end{algorithmic}
\end{algorithm}

DATM proceeds as follows (Algorithm~\ref{alg:otr_quant_kmeans}). First, it randomly selects $K$ tokens from $X$ to form the initial \textit{dst} set, while assigning all remaining tokens to the \textit{src} set \textbf{\textit{(Dst$\&$Src Init.)}}. Next, it computes the pairwise L2 loss between all \textit{dst} and \textit{src} tokens and assigns each \textit{src} token to its nearest \textit{dst} token \textbf{\textit{(Dst–Src Pairing)}}. It then averages the overall token-pair L2 loss and checks whether the change from the previous iteration is smaller than $\epsilon$ \textbf{\textit{(Convergence Check)}}. If the convergence criterion is met, the iterative process ends. Otherwise, it updates each \textit{dst} token by computing the mean of its \textit{dst–src} group and selecting the token closest to this mean \textbf{\textit{(Dst Update)}}, after which the algorithm repeats the sequence of \textit{\textbf{Dst–Src Pairing}}, \textit{\textbf{Convergence Check}}, and \textit{\textbf{Dst Update}} until convergence. Once the iterations finish, the algorithm produces the final \textit{dst–src} index pairs, applies top-k selection based on a predefined ratio, and uses the resulting pairs for token reduction \textit{\textbf{(Top-k Selection)}}. Through this iterative process, our proposed algorithm progressively updates the \textit{dst} tokens to explicitly minimize the token-pair loss, thereby improving the quality of token matching and enabling a higher token reduction ratio.

However, this matching algorithm requires a substantial amount of vector processing due to large-scale loss computations and reduction operations. As a result, it is inefficient on conventional GPU architectures, suffers from excessive latency, and its runtime cannot be fully hidden behind the main diffusion process. These limitations motivate the need for a specialized hardware design tailored to the characteristics of DATM. 

To reduce hardware cost, we apply 4-bit channel-wise quantization to DATM. The resulting accuracy degradation is negligible because the algorithm relies on relative token relationships rather than absolute feature values; small perturbations in individual values have little effect on the similarity ordering among tokens. Furthermore, the iterative refinement process reduces the token-pair loss by a much larger margin than the minor quantization-induced perturbations, effectively compensating for quantization errors and preserving both matching quality and overall performance. In the next section, we describe the hardware architecture that implements ORBIS's diffusion process.

\subsection{ORBIS's Hardware Design}
\begin{figure}[t]
  \centering
  \includegraphics[width=\linewidth]{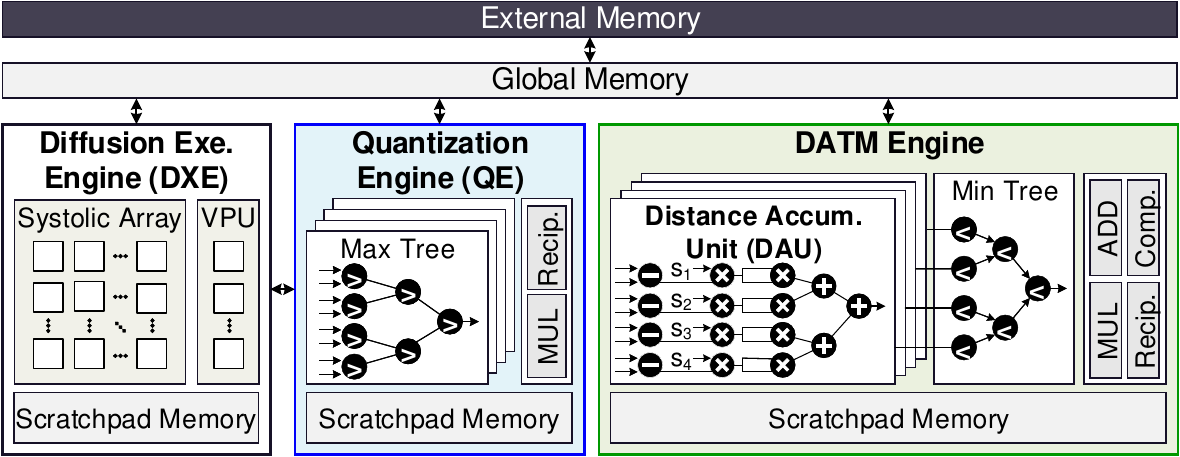}
  \vspace{-0.25in}
  \caption{The overview of ORBIS's hardware architecture}
  \label{fig6}
  \vspace{-0.2in}
\end{figure}

Fig.~\ref{fig6} illustrates the overall ORBIS's hardware architecture, which consists of three main components: Diffusion Execution Engine (DXE), Quantization Engine (QE), and DATM Engine. The DXE performs the main diffusion process and is composed of a Systolic Array (SA) and a Vector Processing Unit (VPU). The QE performs 4-bit channel-wise quantization to produce quantized activations, and the DATM Engine operates on these quantized activations through a deeply pipelined datapath optimized for the vector-processing pattern of the proposed DATM algorithm. These components work together to translate ORBIS’s algorithm-driven high token reduction ratio into real end-to-end speedup by fully hiding the token-matching latency behind the main diffusion process, while incurring minimal hardware overhead.

The ORBIS’s diffusion process proceeds as follows. During the FC timestep, DXE performs the original diffusion process. Once the SDPA output is available, QE computes per-channel scale factors and performs 4-bit quantization using the VPU inside DXE, while SA continues its computation in parallel.
After quantization, the DATM Engine receives the quantized activations and executes the DATM algorithm. The Distance Accumulation Unit (DAU) and Min Tree module jointly perform the \textit{\textbf{Dst–Src Pairing}} stage through a deeply pipelined datapath, applying scale factors, accumulating distances, and performing on-the-fly reductions. This fully pipelined structure drastically shortens DATM latency, allowing it to be completely hidden behind the diffusion computation. The remaining stages—\textit{\textbf{Dst Update}} and \textit{\textbf{Convergence Check}}—are handled by vector units in the DATM engine.
Finally, \textit{\textbf{Top-k Selection}} is performed via a bitonic sorting network~\citep{batcher1968sorting} implemented with comparator arrays. Once the final token index pairs are generated, they are stored in external memory for all layers and directly used for token reduction and restoration during RC timesteps.

\section{Evaluation}
\subsection{Accuracy Evaluation}
% \textbf{Experimental setup.}~To evaluate the accuracy of ORBIS, we employ two representative video diffusion models: 
\textbf{Experimental setup.}~We employ two representative video diffusion models: CogVideoX~\citep{yang2024cogvideox} and HunyuanVideo~\citep{hunyuanvideo2024}. CogVideoX is evaluated on 49-frame videos at a resolution of 480$\times$720, while HunyuanVideo is evaluated on 129-frame videos at a resolution of 544$\times$960. Both models are evaluated using the VBench dataset~\citep{vbench2023}. Video fidelity is measured by the VBench score, and perceptual reconstruction quality is evaluated using PSNR~\citep{psnr2008}, SSIM~\citep{wang2004ssim}, and LPIPS~\citep{zhang2018lpips} relative to the vanilla model’s output. 
% PSNR measures pixel-level reconstruction fidelity, SSIM assesses structural similarity, and LPIPS captures perceptual differences based on deep feature representations. 
For comparison, we benchmark ORBIS against the state-of-the-art algorithm AsymRnR~\citep{sun2025asymrnr}, based on its officially released implementation.

\textbf{Accuracy.}~Table~\ref{table1} reports model accuracy and the average token reduction ratio. Compared with AsymRnR, ORBIS achieves both higher accuracy and a substantially larger reduction ratio. This is enabled by our output-guided and distribution-aware design, which improves token-pair quality to simultaneously attain high accuracy and reduction levels beyond prior methods.

Table~\ref{table2} presents the ablation study on model accuracy and shows that the impact of quantization is negligible. This robustness arises because DATM’s iterative refinement reduces token-pair loss by a margin far exceeding these minor perturbations, effectively neutralizing quantization errors and maintaining matching quality.

\subsection{Performance Evaluation}
\textbf{Experimental setup.}~We implement a custom cycle-level simulator with Ramulator 2.0~\citep{Luo2024ramulator2}. We use the NVIDIA A100 80GB PCIe~\citep{nvidiaA100} as the comparison baseline, and measure its latency and power using NVIDIA Nsight System~\citep{nvidia_nsight_2024} and NVIDIA-SMI~\citep{nvidia_smi}. For a fair comparison, we scale ORBIS’s hardware to 38 instances to match the A100’s FP16 peak compute performance and memory bandwidth, and set its global memory to 40 MB to match the A100’s L2 cache capacity~\citep{nvidia_ampere}. To measure the area and power of ORBIS's hardware, we implement the design in SystemVerilog and synthesize it using Synopsys Design Compiler~\citep{synopsysDC} under a 28nm technology node targeting 1GHz. For energy estimation, we assume a dram energy cost of 3.9 pJ/bit~\citep{OConnor2017FineGrainedDRAM}.

\begin{table}[t]
\caption{Evaluation of model accuracy}
\vspace{-0.1in}
\includegraphics[width=\linewidth]{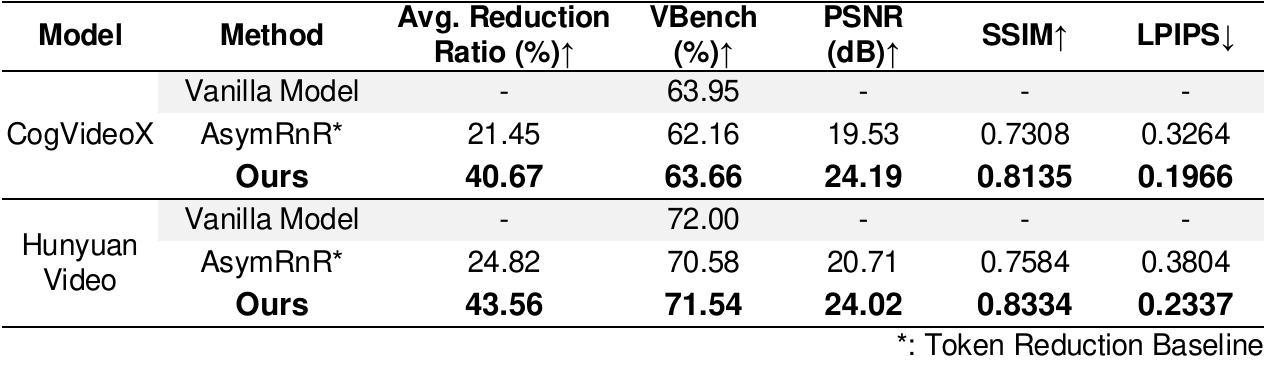}
\label{table1}
\vspace{-0.25in}
\end{table}
\begin{table}[t]
\caption{Ablation study on model accuracy}
\vspace{-0.1in}
\includegraphics[width=\linewidth]{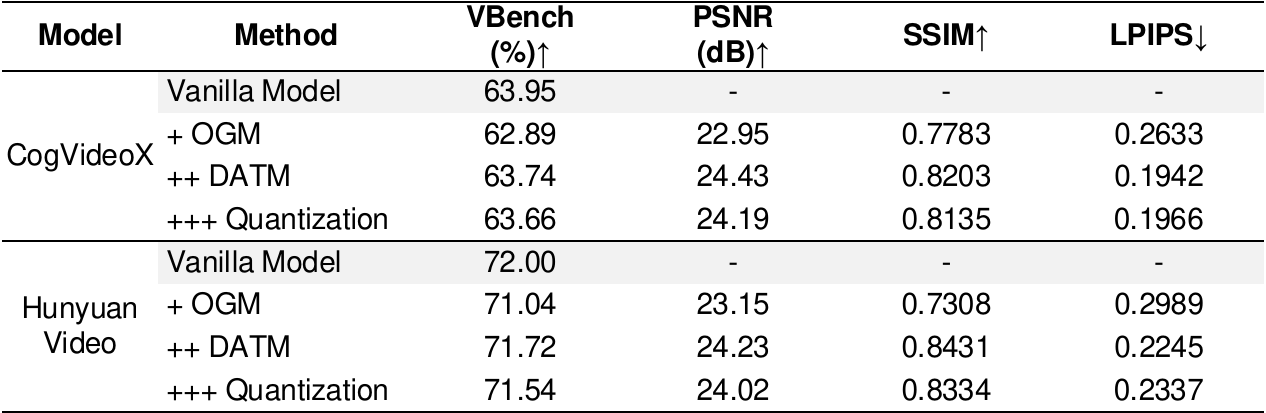}
\label{table2}
\vspace{-0.2in}
\end{table}

\textbf{Configurations.}~We compare ORBIS with the state-of-the-art token reduction approach AsymRnR (on A100 GPU) and evaluate four configurations: ORBIS$_{BASE}$, ORBIS$_{OGM}$, ORBIS$_{OGM+DATM}$, and ORBIS$_{ALL}$.
On the hardware side, the first three configurations use only the Diffusion Execution Engine (DXE), whereas ORBIS$_{ALL}$ uses the full ORBIS's hardware, including the Quantization Engine (QE) and the DATM engine. 
On the software side, ORBIS$_{BASE}$ runs the vanilla model; ORBIS$_{OGM}$ applies only OGM with conventional input-guided matching; ORBIS$_{OGM+DATM}$ adds the non-quantized DATM algorithm; Finally, ORBIS$_{ALL}$ applies all proposed software techniques\textemdash including the quantized DATM algorithm.

\begin{figure}[t]
  \centering
  \includegraphics[width=\linewidth]{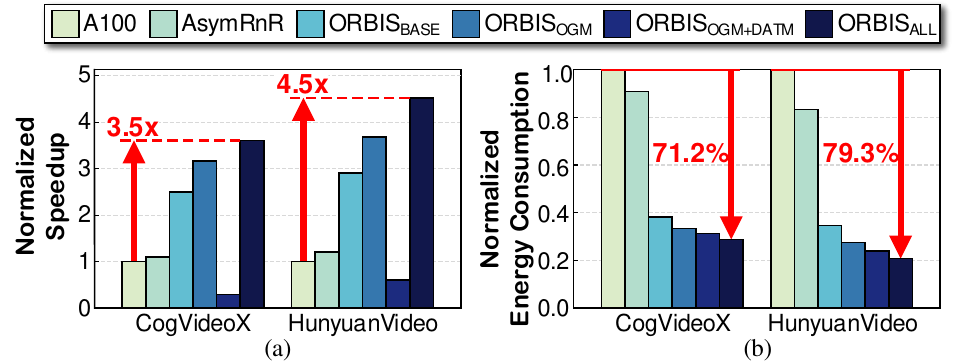}
  \vspace{-0.25in}
  \caption{Evaluation of normalized (a) Speedup, (b) Energy consumption}
  \vspace{-0.15in}
  \label{fig7}
\end{figure}

\textbf{Speedup.}~Fig.~\ref{fig7} (a) shows the normalized speedup of the two models relative to the A100 GPU. AsymRnR achieves only 1.1$\times$–1.2$\times$ speedup due to its limited token reduction ratio. ORBIS$_{BASE}$ reaches 2.7$\times$–2.9$\times$ thanks to ORBIS’s hardware specialization for diffusion workloads. Incorporating OGM further improves the speedup to 3.2$\times$–3.7$\times$ in ORBIS$_{OGM}$. In contrast, ORBIS$_{OGM+DATM}$ drops to 0.3$\times$–0.6$\times$ as the DXE alone lacks the vector-processing capability required by DATM. Finally, ORBIS$_{ALL}$ achieves a 3.5$\times$–4.5$\times$ speedup by leveraging the specialized DATM engine with a deeply pipelined datapath, clearly demonstrating the necessity of hardware support.

\begin{figure}[t]
  \centering
  \includegraphics[width=\linewidth]{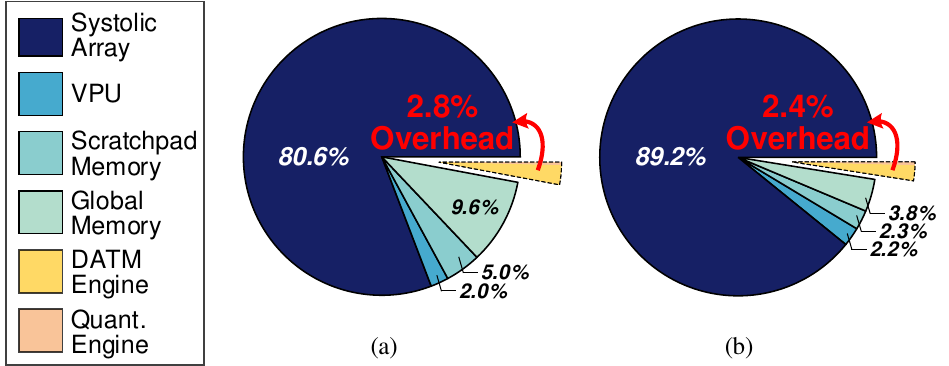}
  \vspace{-0.25in}
  \caption{(a) Area, (b) Power breakdown of ORBIS's hardware}
  \label{fig8}
  \vspace{-0.1in}
\end{figure}

\textbf{Energy Consumption.}~Fig.~\ref{fig7}(b) shows the normalized energy consumption relative to the A100 GPU. AsymRnR achieves only a 9.1\%–16.7\% reduction in energy. ORBIS$_{BASE}$ attains a 61.7\%–65.3\% reduction thanks to the specialized ORBIS hardware, and adding OGM further improves the savings to 66.6\%–72.5\% in ORBIS$_{OGM}$. Notably, ORBIS$_{OGM+DATM}$ attains 68.7\%–76.0\% despite its low speed-up, since energy is determined by total computation rather than latency.
Finally, ORBIS$_{ALL}$ reaches the highest 71.2\%–79.3\%, as running quantized DATM on the specialized engine further decreases energy consumption.

\textbf{Area and power breakdown.}~Fig.~\ref{fig8} summarizes the area and power breakdown of ORBIS’s hardware, which occupies 470.9mm$^{2}$/ 332.1W in 28nm process and scales to 14.8mm$^{2}$/124.5W at 7nm~\citep{Sarangi2021DeepScaleTool}. The corresponding 7nm A100 GPU measures 596.8mm$^{2}$ and 300W, indicating that ORBIS's hardware offers substantially higher area and power efficiency. The systolic array accounts for the majority of resources, while the additional hardware introduced for DATM \textemdash the DATM engine and the quantization engine\textemdash occupies only a negligible fraction of the total area and power. This minimal overhead results from our 4-bit quantization scheme, which greatly reduces the size of each processing unit and enables the DATM algorithm to be implemented with minimal hardware cost.

\section{Conclusion}
ORBIS combines an output-guided token reduction approach and the Distribution-Aware Token Matching algorithm with specialized hardware that fully hides matching latency while incurring minimal area overhead and negligible accuracy degradation. ORBIS achieves almost 2$\times$ higher token reduction ratio than the state-of-the-art method, AsymRnR, and delivers up to 4.5$\times$ speedup and 79.3\% energy reduction compared to an NVIDIA A100 GPU.

\section{Acknowledgement}
This work was supported by the Institute of Information \& Communications Technology Planning \& Evaluation (IITP) through the Graduate School of AI Semiconductor(IITP-2026-RS-2023-00256472), and the IITP-ITRC program(IITP-2026-RS-2020-II201847), all funded by the Korea government(MSIT). The EDA tool was supported by the IC Design Education Center(IDEC), Korea.

\newpage
\bibliographystyle{ACM-Reference-Format}
\bibliography{99_reference}
\footnotesize
\end{document}